\documentclass[11pt,a4paper]{article}
\usepackage[hyperref]{acl2020}
\usepackage{times}
\usepackage{latexsym}
\usepackage{bm}
\usepackage{booktabs}
\usepackage{multirow}

\usepackage{graphicx} 

\usepackage{microtype}
\usepackage{enumitem}
\usepackage{todonotes}

\aclfinalcopy 
\def\aclpaperid{3067} 

\newcommand{\Ni}{({\em i})~}
\newcommand{\Nii}{({\em ii})~}
\newcommand{\Niii}{({\em iii})~}
\newcommand{\Niv}{({\em iv})~}

\title{{W}hat {W}as {W}ritten vs. {W}ho {R}ead {I}t:\\ News Media Profiling Using Text Analysis and Social Media Context}

\author{
    Ramy Baly$^1$,
    Georgi Karadzhov$^2$,
    Jisun An$^3$,
    Haewoon Kwak$^3$,
    Yoan Dinkov$^4$,\\
    \bf Ahmed Ali$^3$,
    James Glass$^1$,
    Preslav Nakov$^3$\\
    $^1$MIT Computer Science and Artificial Intelligence Laboratory\\
    $^2$University of Cambridge,
    $^3$Qatar Computing Research Institute, HBKU,
    $^4$Sofia University\\
    \texttt{\{baly,glass\}@mit.edu},
    \texttt{georgi.karadzhov@cl.cam.ac.uk}\\
    \texttt{\{jan,hkwak,amali,pnakov\}@hbku.edu.qa},
    \texttt{jdinkov@uni-sofia.bg}
}

\date{}

\begin{document}\maketitle

\begin{abstract}

Predicting the political bias and the factuality of reporting of entire news outlets are critical elements of media profiling, which is an understudied but an increasingly important research direction. The present level of proliferation of fake, biased, and propagandistic content online, has made it impossible to fact-check every single suspicious claim, either manually or automatically. Alternatively, we can profile entire news outlets and look for those that are likely to publish fake or biased content. This approach makes it possible to detect likely ``fake news'' the moment they are published, by simply checking the reliability of their source.

From a practical perspective, political bias and factuality of reporting have a linguistic aspect but also a social context. Here, we study the impact of both, namely \Ni~\textit{what was written} (i.e.,~ what was published by the target medium, and how it describes itself on Twitter) vs. \Nii~\textit{who read it} (i.e.,~analyzing the readers of the target medium on Facebook, Twitter, and YouTube). We further study \Niii~\textit{what was written about the target medium} on Wikipedia. The evaluation results show that \textit{what was written} matters most, and that putting all information sources together yields huge improvements over the current state-of-the-art.

\end{abstract}

\section{Introduction}

The rise of the Web has made it possible for anybody to create a website or a blog and to become a \textit{news medium}.
Undoubtedly, this was a hugely positive development as it elevated freedom of expression to a whole new level, thus allowing anybody to make their voice heard online.
With the subsequent rise of social media, anybody could potentially reach out to a vast audience, something that until recently was only possible for major news outlets.

\noindent One of the consequences was a \textit{trust crisis}: with traditional news media stripped off their gate-keeping role, the society was left unprotected against potential manipulation.
The issue became a general concern in 2016, a year marked by micro-targeted online disinformation and misinformation at an unprecedented scale, primarily in connection to Brexit and the US Presidential campaign.
These developments gave rise to the term ``fake news'', which can be defined as ``false, often sensational, information disseminated under the guise of news reporting.''\footnote{\url{www.collinsdictionary.com/dictionary/english/fake-news}}
It was declared Word of the Year 2016 by Macquarie Dictionary and of Year 2017 by the Collins English Dictionary.

In an attempt to solve the trust problem, several initiatives such as Politifact, Snopes, FactCheck, and Full Fact, have been launched to fact-check suspicious claims manually.
However, given the scale of the proliferation of false information online, it became clear that it was unfeasible to fact-check every single suspicious claim, even when this was done automatically, not only due to computational challenges but also due to timing.
In order to fact-check a claim, be it manually or automatically, one often needs to verify the stance of mainstream media concerning that claim and the reaction of users on social media.
Accumulating this kind of evidence takes time, but time flies very fast, and any delay means more potential sharing of the malicious content on social media.
A study has shown that for some very viral claims, more than 50\% of the sharing happens within the first ten minutes after posting the micro-post on social media~\cite{zaman2014}, and thus timing is of utmost importance.  Moreover, an extensive recent study has found that ``fake news'' spreads six times faster and reaches much farther than real news~\cite{Vosoughi1146}.

\noindent A much more promising alternative is to focus on the source and to profile the medium that initially published the news article.
The idea is that media that have published fake or biased content in the past are more likely to do so in the future.
Thus, profiling media in advance makes it possible to detect likely ``fake news'' the moment it is published, by simply checking the reliability of its source.

From a practical perspective, political bias and factuality of reporting have not only a linguistic aspect but also a social context.
Here, we study the impact of both, namely
    \Ni~\textit{what was written} (the text of the articles published by the target medium, the text and the audio signal in the videos of its YouTube channel, as well as how the medium self-describes itself on Twitter) vs.
    \Nii~\textit{who read it} (by analyzing the media readers in Facebook, Twitter, and YouTube). We further study
    \Niii~\textit{what was written about the target medium} on Wikipedia.

\noindent Our contributions can be summarized as follows:
\begin{itemize}[leftmargin=*]
    \item We model the leading political ideology (left, center or right bias) and the factuality of reporting (high, mixed, or low) of news media by modeling the textual content of what they publish vs. who reads it in social media (Twitter, Facebook, and YouTube). The latter is novel for these tasks.
    \item We combine a variety of information sources about the target medium, many of which have not been explored for our tasks, e.g.,~YouTube video channels, political bias estimates of their Facebook audience, and information from the profiles of the media followers on Twitter.
    \item We use features from different data modalities: text, metadata, and speech.  The latter two are novel for these tasks.
    \item We achieve sizeable improvements over the current state-of-the-art for both tasks. 
    \item We propose various ensembles to combine the different types of features, achieving further improvements, especially for bias detection.
    \item We release the data, the features, and the code necessary to replicate our results.
\end{itemize}

In the rest of this paper, we discuss some related work, followed by a description of our system's architecture and the information sources we use.
Then, we present the dataset, the experimental setup, and the evaluation results.
Finally, we conclude with possible directions for future work.

\section{Related Work}\label{sec:related}

While leveraging social information and temporal structure to predict the factuality of reporting of a news medium is not new~\cite{Canini:2011,Castillo:2011:ICT:1963405.1963500,Ma:2015:DRU,ma2016detecting,PlosONE:2016}, modeling this at the medium level is a mostly unexplored problem.
A popular approach to predict the factuality of a medium is to check the general stance of that medium concerning already fact-checked claims~\cite{mukherjee2015leveraging,Popat:2017:TLE:3041021.3055133,Popat:2018:CCL:3184558.3186967}.
Therefore, stance detection became an essential component in fact-checking systems~\cite{baly2018integrating}.

In political science, media profiling is essential for understanding media choice~\cite{iyengar2009red}, voting behavior~\cite{dellavigna2007fox}, and polarization~\cite{graber2017mass}.
The outlet-level bias is measured as a similarity of the language used in news media to political speeches of congressional Republicans or Democrats, also used to measure media slant~\cite{gentzkow2006media}.
Article-level bias was also measured via crowd-sourcing~\cite{budak2016fair}.
Nevertheless, public awareness of media bias is limited~\cite{elejalde2018nature}.

Political bias was traditionally used as a feature for fact verification~\cite{DBLP:journals/corr/abs-1803-10124}.
In terms of modeling, \citet{Horne:2018:ANL:3184558.3186987} focused on predicting whether an article is biased or not.
Political bias prediction was explored by \citet{DBLP:journals/corr/PotthastKRBS17} and \citet{saleh2019team}, where news articles were modeled as left vs. right, or as hyperpartisan vs. mainstream.
Similarly, \citet{kulkarni2018multi} explored the left vs. right bias at the article level, modeling both textual and URL contents of articles.

In our earlier research~\citep{baly2018predicting}, we analyzed both the political bias and the factuality of news media.
We extracted features from several sources of information, including articles published by each medium, what is said about it on Wikipedia, metadata from its Twitter profile, in addition to some web features (URL structure and traffic information).
The experiments on the Media Bias/Fact Check (MBFC) dataset showed that combining features from these different sources of information was beneficial for the final classification.
Here, we expand this work by extracting new features from the existing sources of information, as well as by introducing new sources, mostly related to the social media context, thus achieving sizable improvements on the same dataset.

\begin{figure*}[tbh]
    \centering
    \includegraphics[width=1\textwidth]{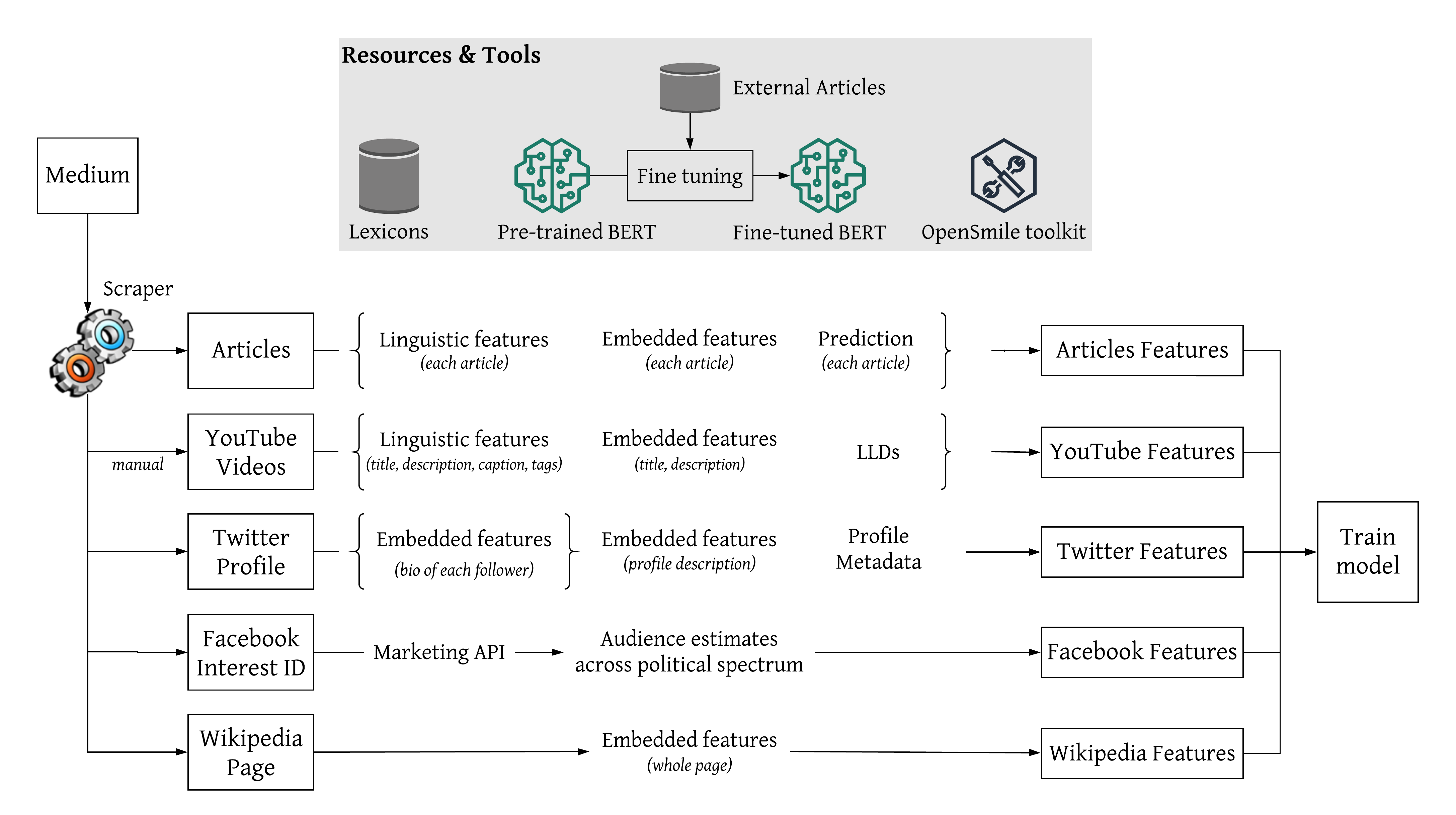}
    \caption{The architecture of our system for predicting the political bias and the factuality of reporting of news media.
    The features inside $\{$curly brackets$\}$ are calculated at a finer level of granularity and are then aggregated at the medium level.
    The upper gray box shows the resources used to generate features, e.g.,~the OpenSmile toolkit is used to extract low-level descriptors (LLD) from YouTube videos;
    see Section~\ref{sec:models_features} for further details.}
    \label{fig:architecture}
\end{figure*}

\noindent In follow-up work~\citep{baly2019multi}, we showed that jointly predicting the political bias and the factuality is beneficial, compared to predicting each of them independently.
We used the same sources of information as in~\citep{baly2018predicting}, but the results were slightly lower. While here we focus on analyzing political bias and factuality separately, future work may analyze how the newly proposed features and sources affect the joint prediction.

\section{System and Features}\label{sec:models_features}

In this section, we present our system. For each target medium, it extracts a variety of features to model
    \Ni what was written by the medium,
    \Nii the audience of the medium on social media, and
    \Niii what was written about the medium in Wikipedia.
This results in multi-modal (text, speech, and metadata) feature set, which we use to train a classifier to predict the political bias and the factuality of reporting of news media.
Figure~\ref{fig:architecture} illustrates the system architecture.

\subsection{What Was Written}
We describe the features that we used to model the content generated by the news media, analyzing both the articles they publish on their website as well as relevant activity on social media.

\subsubsection{Articles on the News Medium Website}
Given a target news medium, we first collect a number of articles it has published. Then, we extract various types of features from the text of these articles. Below we describe these features in more detail.

\textit{Linguistic Features:}
These features focus on language use, and they model text structure, topic, sentiment, subjectivity, complexity, bias, and morality.
They have proved useful for detecting fake articles, as well as for predicting the political bias and the factuality of reporting of news media~\cite{DBLP:journals/corr/abs-1803-10124,baly2018predicting}.
We extracted such features using the News Landscape (NELA) toolkit~\cite{DBLP:journals/corr/abs-1803-10124}, and we will refer to them as the NELA features in the rest of this paper.
We averaged the NELA features for the individual articles in order to obtain a NELA representation for a news medium.
Using arithmetic averaging is a good idea as it captures the general trend of articles in a medium, while limiting the impact of outliers.
For instance, if a medium is known to align with left-wing ideology, this should not change if it published a few articles that align with right-wing ideology.
We use this method to aggregate all features that we collected at a level of granularity that is finer than the medium-level.

\textit{Embedding Features:}
We encoded each article using BERT~\cite{devlin2018bert} by feeding the first 510 WordPieces\footnote{There is a limit of maximum of 512 input tokens, and we had to leave space for the special tokens \textsc{[CLS]} and \textsc{[SEP]}.} from the article\footnote{This is recommended in~\cite{adhikari2019docbert} when encoding full documents using Transformer-based models.} and then averaging the word representations extracted from the second-to-last layer.\footnote{This is common practice, since the last layer may be biased towards the pre-training objectives of BERT.}
In order to obtain representations that are relevant to our tasks, we fine-tuned BERT by training a softmax layer on top of the \textsc{[CLS]} output vector to predict the label (bias or factuality) of news articles that are scrapped from an external list of media to avoid overfitting.
The articles' labels are assumed to be the same as those of the media in which they are published (a form of distant supervision).
This is common practice in tasks such as ``fake news'' detection, where it is difficult to manually annotate large-scale datasets~\cite{norregaard2019nela}.
We averaged the BERT representations across the articles in order to aggregate them at the medium level.

\textit{Aggregated Probabilities:}
We represent each article by a $\mathcal{C}$-dimensional vector that corresponds to its posterior probabilities of belonging to each class $c_i$, $i\in\{1, \dots, \mathcal{C}\}$ of the given task, whether it is predicting the political bias or the factuality of the target news medium.
These probabilities are produced by training a softmax layer on top of the \textsc{[CLS]} token in the above-mentioned fine-tuned BERT model.
We averaged the probability representations across the articles in order to aggregate them at the medium level.

\subsubsection{YouTube Video Channels}

Some news media post their video content on YouTube. Thus, we use YouTube channels by modeling their textual and acoustic contents to predict the political bias and the factuality of reporting of the target news medium.
This source of information is relatively underexplored, but it has demonstrated potential for modeling bias~\citep{INTERSPEECH2019:youtube} and factuality~\cite{ASRU2019:deception}.

Due to the lack of viable methods for automatic channel retrieval, we manually looked up the YouTube channel for each medium.
For each channel marked as English, we crawled 25 videos (on average) with at least 15 seconds of speech content.
Then, we processed the speech segments from each video into 15-second episodes by mapping the duration timeline to the subtitle timestamps.

\noindent We used the OpenSMILE toolkit~\cite{eyben2010opensmile} to extract low-level descriptors (LLDs) from these speech episodes, including frame-based features (e.g., energy), fundamental frequency, and Mel-frequency cepstral coefficients (MFFC). This set of features proved to be useful in the Interspeech Computational Paralinguistics challenge of emotion detection ~\cite{schuller2009interspeech}.
To complement the acoustic information, we retrieved additional textual data such as descriptions, titles, tags, and captions.
This information is encoded using a pre-trained BERT model. Furthermore, we extracted the NELA features from the titles and from the descriptions.
Finally, we averaged the textual and the acoustic features across the videos to aggregate them at the medium level.

\subsubsection{Media Profiles in Twitter}

We model how news media portray themselves to their audience by extracting features from their Media Twitter profiles.

In our previous work, this has proven useful for political bias prediction~\cite{baly2018predicting}.
Such features include information about whether Twitter verified the account, the year it was created, its geographical location, as well as some other statistics, e.g., the number of followers and of tweets posted.

We encoded the profile's description using SBERT for the following reasons:
    \Ni~unlike the articles, the number of media profiles is too small to fine-tune BERT, and
    \Nii~most Twitter descriptions have sentence-like structure and length.
If a medium has no Twitter account, we used a vector of zeros.

\subsection{Who Read it}

We argue that the audience of a news medium can be indicative of the political orientation of that medium.
We thus propose a number of features to model this, which we describe below.

\subsubsection{Twitter Followers Bio}

Previous research has used the followers' networks and the retweeting behavior in order to infer the political bias of news media~\cite{wong2013quantifying,CoNLL2019:troll:roles,ICWSM2020:Unsupervised:Stance:Twitter}.
Here, we analyze the self-description (bio) of Twitter users that follow the target news medium.
The assumption is that (\emph{i})~followers would likely agree with the news medium's bias, and (\emph{ii})~they might express their own bias in their self-description.

\noindent We retrieved the public profiles of 5,000 followers for each target news medium with a Twitter account, and we excluded those with non-English bios since our dataset is mostly about US media.
Then, we encoded each follower's bio using SBERT~\cite{reimers2019sentence}.
As we had plenty of followers' bios, this time fine-tuning BERT would have been feasible. However, we were afraid to use distant supervision for labeling as we did with the articles since people sometimes follow media with different political ideologies.
Thus, we opted for SBERT, and we averaged the SBERT representations across the bios in order to obtain a medium-level representation.

\subsubsection{Facebook Audience}

Like many other social media giants, Facebook makes its revenues from advertisements.
The extensive user interaction enables Facebook to create detailed profiles of its users, including demographic attributes such as age, gender, income, and political leaning.
Advertisers can explore these attributes to figure out the targeting criteria for their ads, and Facebook returns an audience estimate based on these criteria.
For example, the estimated number of users who are female, 20-years-old, very liberal, and interested in the NY Times is 160K.
These estimates have been used as a proxy to measure the online population in various domains~\cite{fatehkia2018using,araujo2017using,ribeiro2018media}.

In this study, we explore the use of political leaning estimates of users who are interested in particular news media.
To obtain the audience estimates for a medium, we identify its Interest ID using the Facebook Marketing API
\footnote{\url{http://developers.facebook.com/docs/marketing-api}}.
Given an ID, we retrieve the estimates of the audience (in the United States) who showed interest in the corresponding medium.
Then, we extract the audience distribution over the political spectrum, which is categorized into five classes ranging from \textit{very conservative} to \textit{very liberal}.

\subsubsection{YouTube Audience Statistics}

Finally, we incorporate audience information from YouTube videos.
We retrieved the following metadata to model audience interaction: number of views, likes, dislikes, and comments for each video.
As before, we averaged these statistics across the videos to obtain a medium-level representation.

\subsection{What Was Written About the Target Medium}

Wikipedia contents describing news media were useful for predicting the political bias and the factuality of these media~\cite{baly2018predicting}.
We automatically retrieved the Wikipedia page for each medium, and we encoded its contents using the pre-trained BERT model.\footnote{Similarly to Twitter descriptions, the number of news media with Wikipedia pages is too small to fine-tune BERT.}
Similarly to encoding the articles, we fed the encoder with the first 510 tokens of the page's content, and used as an output representation the average of the word representations extracted from the second-to-last layer.
If a medium had no page in Wikipedia, we used a vector of zeros.

\section{Experiments and Evaluation}
\label{sec:experiments}

\subsection{Dataset}

We used the Media Bias/Fact Check (MBFC) dataset, which consists of a list of news media along with their labels of both political bias and factuality of reporting.
Factuality is modeled on a 3-point scale: \textit{low}, \textit{mixed}, and \textit{high}.
Political bias is modeled on a 7-point scale: \textit{extreme-left}, \textit{left}, \textit{center-left}, \textit{center}, \textit{center-right}, \textit{right}, and \textit{extreme-right}.
Further details and examples of the dataset can be found in~\cite{baly2018predicting}.

After manual inspection, we noticed that the \textit{left-center} and \textit{right-center} labels are ill-defined, ambiguous transitionary categories.
Therefore, we decided to exclude news media with these labels.
Also, to reduce the impact of potentially subjective decisions made by the annotators, we merged the \textit{extreme-left} and \textit{extreme-right} media with the left and right categories, respectively.
As a result, we model political bias on a 3-point scale (\textit{left}, \textit{center}, and \textit{right}), and the dataset got reduced to 864 news media.
Table~\ref{tab:stats} provides statistics about the dataset.

\begin{table}[ht]
\centering
\begin{tabular}{lllr}
\toprule
\multicolumn{2}{c}{\bf Political Bias} & \multicolumn{2}{c}{\bf Factuality} \\
\midrule
Left    &   243 &   Low     &   162\\
Center  &   272 &   Mixed   &   249\\
Right   &   349 &   High    &   453\\
\bottomrule
\end{tabular}
\caption{Label counts in the dataset.\label{tab:stats}}
\end{table}

We were able to retrieve Wikipedia pages for 61.2\% of the media, Twitter profiles for 72.5\% of the media, Facebook pages for 60.8\% of the media, and YouTube channel for 49\% of the media.

\subsection{Experimental Setup}

We evaluated the following aspects about news media separately and in combinations: \Ni~what the target medium wrote, \Nii~who read it, and \Niii~what was written about that medium.
We used the features described in Section~\ref{sec:models_features} to train SVM classifiers for predicting the political bias and the factuality of reporting of news media.
We performed an incremental ablation study by combining the best feature(s) from each aspect to obtain a combination that achieves even better results.
We used 5-fold cross-validation to train and to evaluate an SVM model using different features and feature combinations.
At each iteration of the cross-validation, we performed a grid search to tune the hyper-parameters of our SVM model, namely the values of the cost $C$ and of the $\gamma$ value for the RBF kernel. In the process of search, we optimized for macro-average $F_1$ score, i.e.,~averaging over the classes, since our dataset is not balanced, which is true for both tasks.
Finally, we evaluated the model on the remaining unseen fold. Ultimately, we report both macro-$F_1$ score, and accuracy.

We compared our results to the majority class baseline and to our previous work~\cite{baly2018predicting}.
The latter used
\Ni~NELA features from articles,
\Nii~embedding representations of Wikipedia pages using averaged GloVe word embeddings,
\Niii~metadata from the media's Twitter profiles, and
\Niv~URL structural features.
Since we slightly modified the MBFC dataset, we retrained the old model on the new version of the dataset.\footnote{The data and the corresponding code, both old and new, are available at \url{https://github.com/ramybaly/News-Media-Reliability}}

To fine-tune BERT's weights, we trained a softmax layer on top of the \textsc{[CLS]} token of the pre-trained BERT model to classify articles for the task at hand: either predicting the articles' political bias as \textit{left}, \textit{center}, or \textit{right}, or predicting their level of factuality as \textit{low} or \textit{high}.\footnote{We ignored \textit{mixed} as it does not apply to articles.}
To avoid overfitting, we scrapped articles from news media listed in the Media Bias/Fact Check database, but not included in our dataset:
30K articles from 298 such media.

Finally, we used two strategies to evaluate feature combinations.
The first one trains a single classifier using all features.
The second one trains a separate classifier for each feature type and then uses an ensemble by taking a weighted average of the posterior probabilities of the individual models.

\noindent Note that we learn different weights for the different models, which ensures that we pay more attention to the probabilities produced by better models.
We used the sklearn library to obtain probabilities from an SVM classifier as a function of the distance between the data point and the learned hyperplane using Platt scaling (for the binary case) or an extension thereof (for the 3-way case).

\begin{table*}[tbh]
\centering
\scalebox{0.8}{
\begin{tabular}{lclrcc}
\toprule
  \bf Group & \bf \#  & {\bf Features} & {\bf Dim.} & {\bf Macro $\boldsymbol{F_1}$} & {\bf Accuracy} \\
\midrule
\multirow{2}{*}{\textit{Baselines}}
 & 1  & Majority class                                              & --    &  19.18  &  40.39 \\
 & 2  & Best model from \cite{baly2018predicting}                   & 764   &  72.90  &  73.61 \\
\midrule
 & 3  & Articles: NELA                                              & 141   &  64.82  &  68.18 \\
 & 4  & Articles: BERT representations                                   & 768   &  79.34  &  79.75 \\
 & 5  & Articles: BERT probabilities                                & 3     &  61.21  &  62.27 \\
 & 6  & Twitter Profiles: Sentence BERT                             & 768   &  59.23  &  60.88 \\
 & 7  & YouTube: NELA (title, description)                          & 260   &  45.78  &  50.46 \\
 & 8  & YouTube: OpenSmile (LLDs)                                   & 385   &  46.13  &  50.69 \\
$A$. \textit{What}
 & 9 & YouTube: BERT (title, description, tags)                    & 768   &  48.36  &  53.94 \\
\textit{Was Written}
 & 10 & YouTube: BERT (captions)                                    & 768   &  49.14  &  53.94 \\ \cmidrule{2-6}
 & 11 & Articles: ALL \textit{(c)}                                   & 912   &  81.00  &  81.48 \\
 & 12 & \bf Articles: ALL \textit{(en)}                              & 912   &  \bf 81.27  &  \bf 81.83 \\
 & 13 & Articles $+$ Twitter Prof. \textit{(c)}                      & 1,691 &  76.59  &  77.20 \\
 & 14 & Articles $+$ Twitter Prof. \textit{(en)}                     & 1,691 &  80.00  &  80.56 \\
 & 15 & Articles $+$ Twitter Prof. $+$ YouTube cap. \textit{(c)}     & 2,315 &  75.73  &  76.39 \\
 & 16 & Articles $+$ Twitter Prof. $+$ YouTube cap. \textit{(en)}    & 2,315 &  79.70  &  80.32 \\
\midrule
 & 17 & Twitter Follower: Sentence BERT                             & 768   &  62.85  &  65.39 \\
 & 18 & YouTube: Metadata                                           & 5     &  40.05  &  46.53 \\
 & 19 & Facebook: Political Leaning Estimates                       & 6     &  27.87  &  43.87 \\ \cmidrule{2-6}
$B$. \textit{Who}
 & 20 & Twitter Fol. $+$ YouTube Meta. \textit{(c)}                  & 773   &  63.72  &  65.86 \\
\textit{Read It}
 & 21 & \bf Twitter Fol. $+$ YouTube Meta. \textit{(en)}             & 773   &  \bf 65.12  &  \bf 66.44 \\
 & 22 & Twitter Fol. $+$ YouTube Meta. $+$ Facebook Estimates \textit{(c)}   & 779   &  63.63  &  65.74 \\
 & 23 & Twitter Fol. $+$ YouTube Meta. $+$ Facebook Estimates \textit{(en)}  & 779   &  64.18  &  66.20 \\
\midrule
$C$. \textit{What} & & & & & \\
\textit{Was Written}
 & 24 & \bf Wikipedia: BERT                                         & 768   &  \bf 64.36  &  \bf 66.09 \\
\textit{About the Medium} & & & & & \\
\midrule
\multirow{8}{*}{\textit{Combinations}}
 & 25 & All features: rows 3--11; 18--20; 25 \textit{(c)}            & 5,413 &  78.17  &  78.70 \\
 & 26 & All features: rows 3--11; 18--20; 25 \textit{(en)}           & 5,413 &  79.42  &  80.32 \\
 & 27 & A$+$B: rows 12 \& 21 \textit{(c)}                            & 1,685 &  84.28  &  84.87 \\ 
 & 28 & A$+$B: rows 12 \& 21 \textit{(en)}                           & 1,685 &  84.15  &  84.64 \\ 
 & 29 & A$+$C: rows 12 \& 24 \textit{(c)}                            & 1,680 &  81.53  &  81.98 \\ 
 & 30 & A$+$C: rows 12 \& 24 \textit{(en)}                           & 1,680 &  82.99  &  83.48 \\ 
 & 31 & A$+$B$+$C: rows 12, 21 \& 24 \textit{(c)}                    & 1,691 &  83.53  &  84.02 \\ 
 & 32 & \bf A$+$B$+$C: rows 12, 21 \& 24 \textit{(en)}               & 1,691 &  \bf 84.77  &  \bf 85.29 \\
\bottomrule
\end{tabular}}
\caption{\textbf{Political bias prediction:} ablation study of the proposed features.
\textit{Dim} refers to the number of features, whereas \textit{(c)} and \textit{(en)} indicate whether the features are concatenated or an ensemble was used, respectively.\label{tab:results_bias}}
\end{table*}

\subsection{Political Bias Prediction}

Table~\ref{tab:results_bias} shows the evaluation results for political bias prediction, grouped according to different aspects. For each aspect, the upper rows correspond to individual features, while the lower ones show combinations thereof.

The results in rows~3--5 show that averaging embeddings from a fine-tuned BERT to encode articles (row~4) works better than using NELA features (row~3).
They also show that using the posterior probabilities obtained from applying a softmax on top of BERT's \textsc{[CLS]} token (row~5) performs worse than using average embeddings (row~4).
This suggest that it is better to incorporate information from the articles' word representations rather than using \textsc{[CLS]} as a compact representation of the articles.
Also, since our BERT was fine-tuned on articles with noisy labels obtained using distant supervision, its predictions for individual articles are also noisy, and so are the vectors of posterior. Yet, this fine-tuning seems to yield improved article-level representations for our task.

The results in rows 7--10 show that captions are the most useful type of feature among those extracted from YouTube.
This makes sense since captions contain the most essential information about the contents of a video.
We can further see that the BERT-based features outperform the NELA ones.
Overall, the YouTube features are under-performing since for half of the media we could not find a corresponding YouTube channel, and we used representations containing only zeroes.

Rows 11-16 show the results for systems that combine article, Twitter, and YouTube features, either directly or in an ensemble.
We can see on rows 13--16 that the YouTube and the Twitter profile features yield loss in performance when added to the article features (rows~11--12). Note that the article features already outperform the individual feature types from rows 3--10 by a wide margin, and thus we will use them to represent the \textit{What Was Written} aspect of the model in our later experiments below.

\noindent We can further notice that the ensembles consistently outperform feature concatenation models, which is actually true for all feature combinations in Table~\ref{tab:results_bias}.

\begin{table*}[tbh]
\centering
\scalebox{0.85}{
\begin{tabular}{lclrcc}
\toprule
  \bf Group & \bf \#  & {\bf Features} & {\bf Dim.} & {\bf Macro $\boldsymbol{F_1}$} & {\bf Accuracy} \\
\midrule
\multirow{2}{*}{\textit{Baselines}}
 & 1  & Majority class                                              & --    &  22.93  &  52.43 \\
 & 2  & Best model from \cite{baly2018predicting}                   & 764   &  61.08  &  66.45 \\
\midrule
 & 3  & Articles: NELA                                              & 141   &  55.54  &  62.62 \\
 & 4  & Articles: BERT representations                                   & 768   &  61.46  &  67.94 \\
 & 5  & Articles: BERT probabilities                                & 3     &  51.39  &  61.46 \\
 & 6  & Twitter Profiles: Sentence BERT                             & 768   &  49.96  &  56.71 \\
 & 7  & YouTube: NELA (title, description)                          & 260   &  32.52  &  51.04 \\
 & 8  & YouTube: OpenSmile (LLDs)                                   & 385   &  37.17  &  52.08 \\
$A$. \textit{What}
 & 9  & YouTube: BERT (title, description, tags)                    & 768   &  38.19  &  54.28 \\
\textit{Was Written}
 & 10 & YouTube: BERT (captions)                                    & 768   &  38.82  &  55.56 \\ \cmidrule{2-6}
 & 11 & Articles: ALL \textit{(c)}                                   & 912   &  59.34  &  64.82 \\
 & 12 & Articles: ALL \textit{(en)}                                  & 912   &  48.27  &  59.95 \\
 & 13 & Articles: BERT $+$ Twitter Prof. \textit{(c)}                & 1,691 &  61.06  &  66.09 \\
 & 14 & \bf Articles: BERT $+$ Twitter Prof. \textit{(en)}           & 1,691 &  \bf 61.50  &  \bf 68.63 \\
 & 15 & Articles: BERT $+$ Twitter Prof. $+$ YouTube: cap. \textit{(c)}     & 2,315 &  60.23  &  65.51 \\
 & 16 & Articles: BERT $+$ Twitter Prof. $+$ YouTube: cap. \textit{(en)}    & 2,315 &  58.21  &  66.44 \\
\midrule
 & 17 & Twitter Follower: Sentence BERT                             & 768   &  42.19  &  58.45 \\
 & 18 & YouTube: Metadata                                           & 5     &  31.92  &  52.78 \\
 & 19 & Facebook: Political Leaning Estimates                       & 6     &  27.24  &  53.70 \\ \cmidrule{2-6}
$B$. \textit{Who}
 & 20 & \bf Twitter Fol. $+$ YouTube Meta. \textit{(c)}              & 773   &  \bf 42.48  &  \bf 58.76 \\
\textit{Read It}
 & 21 & Twitter Fol. $+$ YouTube Meta. \textit{(en)}                 & 773   &  39.66  &  57.64 \\
 & 22 & Twitter Fol. $+$ YouTube Meta. $+$ Facebook Estimates \textit{(c)}   & 779   &  42.28  &  57.76 \\
 & 23 & Twitter Fol. $+$ YouTube Meta. $+$ Facebook Estimates \textit{(en)}  & 779   &  39.33  &  57.99 \\
\midrule
$C$. \textit{What} & & & & & \\
\textit{Was Written}
 & 24 & \bf Wikipedia: BERT                                         & 768   &  \bf 45.74  &  \bf 55.32 \\
\textit{About the Medium} & & & & & \\
\midrule
\multirow{8}{*}{\textit{Combinations}}
 & 25 & All features: rows 3--10; 17--19; 24 \textit{(c)}            & 5,413 &  62.42  &  67.79 \\
 & 26 & All features: rows 3--10; 17--19; 24 \textit{(en)}           & 5,413 &  45.24  &  60.42 \\
 & 27 & A$+$B: rows 14 \& 24 \textit{(c)}                            & 1,680 &  65.45  &  70.40 \\ 
 & 28 & A$+$B: rows 14 \& 24 \textit{(en)}                           & 1,680 &  61.80  &  69.25 \\ 
 & 29 & \bf A$+$C: rows 14 \& 20 \textit{(c)}                        & 1,685 &  \bf 67.25  &  \bf 71.52 \\ 
 & 30 & A$+$C: rows 14 \& 20 \textit{(en)}                           & 1,685 &  62.53  &  69.90 \\ 
 & 31 & A$+$B$+$C: rows 14, 20 \& 24 \textit{(c)}                    & 1,691 &  64.14  &  69.36 \\ 
 & 32 & A$+$B$+$C: rows 14, 20 \& 24 \textit{(en)}                   & 1,691 &  60.35  &  68.90 \\
\bottomrule
\end{tabular}}
\caption{\textbf{Factuality of reporting:} ablation study of the proposed features.
\textit{Dim} refers to the number of features, whereas \textit{(c)} and \textit{(en)} indicate whether the features are concatenated or an ensemble was used, respectively.\label{tab:results_fact}}
\end{table*}

Next, we compare rows 6 and 17, which show results when using Twitter information of different nature: from the target medium profile (row~6) vs. from the profiles of the followers of the target medium (row~17).
We can see that the latter is much more useful, which confirms the importance of the \textit{Who Read It} aspect, which we have introduced in this paper.
Note that here we encode the descriptions and the self-description bio information using Sentence BERT instead of the pre-trained BERT; this is because, in our preliminary experiments (not shown in the table), we found the former to perform much better than the latter.

Next, the results in rows~20--23 show that the YouTube metadata features improve the performance when combined with the Twitter followers' features.
On the other hand, the Facebook audience features' performance is deficient and hurts the overall performance, i.e.,~these estimates seem not to correlate well with the political leanings of news media.
Also, as pointed by~\cite{flaxman2016filter}, social networks can help expose people to different views, and thus the polarization in news readership might not be preserved.

Row~24 shows that the Wikipedia features perform worse than most individual features above, which can be related to coverage as only 61.2\% of the media in our dataset have a Wikipedia page.
Nevertheless, these features are helpful when combined with features about other aspects; see below.

Finally, rows~25--32 show the results when combining all aspects.
The best results are achieved using the best features selected from each of the three aspects in an ensemble setting (row~32).
This combination improves over using information from the article only (row~12) by +3.5 macro-$F_1$ points absolute.
It further yields sizeable absolute improvements over the baseline system from~\cite{baly2018predicting}: by +11.87 macro-$F_1$macro-$F_1$ points.
While a lot of this improvement is due to improved techniques for text representation such as using fine-tuned BERT instead of averaged GloVe word embeddings, modeling the newly-introduced media aspects further yielded a lot of additional improvements.

\subsection{Factuality Prediction}

Table~\ref{tab:results_fact} reports the evaluation results when using the proposed sources/features for the task of predicting the factuality of reporting of news media.

Similarly to the results for political bias prediction, rows 3--10 suggest that the features extracted from articles are more important than those coming from YouTube or from Twitter profiles, and that using BERT to encode the articles yields the best results.
Note that overall, the results in this table are not as high as those for bias prediction.
This reflects the level of difficulty of this task, and the fact that, in order to predict factuality, one needs external information or a knowledge base to be able to verify the published content.

\noindent The results in rows 11--16 show that combining the Twitter profile features with the BERT-encoded articles improves the performance over using the article text only.

Comparing rows 6 and 17 in Table~\ref{tab:results_fact}, we can see that the Twitter follower features perform worse than using Twitter profiles features; this is the opposite of what we observed in Table~\ref{tab:results_bias}.
This makes sense since our main motivation to look at the followers' profiles was to detect political bias, rather than factuality.
Moreover, the metadata collected from media profiles about whether the corresponding account is verified, or its level of activity or connectivity (counts of friends and statuses) are stronger signals for this task.

Finally, rows~25--32 show the results for modeling combinations of the three aspects we are exploring in this paper.
The best results are achieved using the best features selected from the \textit{What was written} and the \textit{What was written about the target medium} aspects, concatenated together.
This combination achieves sizeable improvements compared to the baseline system from~\cite{baly2018predicting}: by +6.17 macro-$F_1$ points absolute.
This result indicates that looking at the audience of the medium is not as helpful for predicting factuality as it was for predicting political bias, and that looking at what was written about the medium on Wikipedia is more important for this task.

\section{Conclusion and Future Work}
\label{sec:future_work}

We have presented experiments in predicting the political ideology, i.e.,~left/center/right bias, and the factuality of reporting, i.e.,~high/mixed/low, of news media.
We compared the textual content of what media publish vs. who read it on social media, i.e.,~on Twitter, Facebook, and YouTube. We further modeled what was written about the target medium in Wikipedia.

We have combined a variety of information sources, many of which were not explored for at least one of the target tasks, e.g.,~YouTube channels, political bias of the Facebook audience, and information from the profiles of the media followers on Twitter.  We further modeled different modalities: text, metadata, and speech signal.  
The evaluation results have shown that while what was written matters most, the social media context is also important as it is complementary, and putting them all together yields sizable improvements over the state of the art.

\noindent In future work, we plan to perform user profiling with respect to polarizing topics such as \emph{gun control}~\cite{ICWSM2020:Unsupervised:Stance:Twitter}, which can then be propagated from users to media~\cite{CoNLL2019:troll:roles,ACL2020:Topical:Stance}. We further want to model the network structure, e.g.,~using graph embeddings~\cite{ICWSM2020:Unsupervised:Stance:Twitter}.
Another research direction is to profile media based on their stance with respect to previously fact-checked claims~\cite{NAACL2018:stance,ACL2020:Known:Lie}, or by the proportion and type of propaganda techniques they use~\cite{EMNLP2019:propaganda:finegrained,IJCAI2020:survey}.
Finally, we plan to experiment with other languages.

\section*{Acknowledgments}
This research is part of the Tanbih project\footnote{\url{http://tanbih.qcri.org/}}, which aims to limit the effect of ``fake news,'' propaganda and media bias by making users aware of what they are reading.
The project is developed in collaboration between the Qatar Computing Research Institute, HBKU and the MIT Computer Science and Artificial Intelligence Laboratory.

\bibliographystyle{acl_natbib}
\bibliography{bibliography}

\end{document}